\documentclass{llncs}
\usepackage{url}

\usepackage{graphicx}
\usepackage{xcolor}
\usepackage{hyperref}
\usepackage[backend=bibtex,url=false,eprint=false,isbn=false, doi=false]{biblatex}
\begin{document}
\mainmatter

\title{Open-ended search for environments and adapted agents using MAP-Elites}
\titlerunning{Open-ended search for environments and adapted agents using MAP-Elites}

\author{Emma Stensby Norstein\inst{1} \and Kai Olav Ellefsen\inst{1} \and Kyrre Glette\inst{1,2}}
\authorrunning{Emma Stensby Norstein, Kai Olav Ellefsen, Kyrre Glette}
\institute{Department of Informatics, University of Oslo, Oslo, Norway \and RITMO, University of Oslo, Oslo, Norway}

\maketitle

\begin{abstract}
Creatures in the real world constantly encounter new and diverse challenges they have never seen before. They will often need to adapt to some of these tasks and solve them in order to survive. This almost endless world of novel challenges is not as common in virtual environments, where artificially evolving agents often have a limited set of tasks to solve. An exception to this is the field of open-endedness where the goal is to create unbounded exploration of interesting artefacts. We want to move one step closer to creating simulated environments similar to the diverse real world, where agents can both find solvable tasks, and adapt to them. Through the use of MAP-Elites we create a structured repertoire, a map, of terrains and virtual creatures that locomote through them. By using novelty as a dimension in the grid, the map can continuously develop to encourage exploration of new environments. The agents must adapt to the environments found, but can also search for environments within each cell of the grid to find the one that best fits their set of skills. Our approach combines the structure of MAP-Elites, which can allow the virtual creatures to use adjacent cells as stepping stones to solve increasingly difficult environments, with open-ended innovation. This leads to a search that is unbounded, but still has a clear structure. We find that while handcrafted bounded dimensions for the map lead to quicker exploration of a large set of environments, both the bounded and unbounded approach manage to solve a diverse set of terrains.

\keywords{Evolutionary algorithms, Virtual creatures, Environments, Map-Elites, Open-endedness, Modular robots}
\end{abstract}

\section{Introduction}

Virtual creatures that \emph{learn} locomotion skills have attracted significant research interest. 
Even so there has not been much research that combines the optimisation of the controller, morphology and environment. All three of these components play an important role in determining the behaviour of an agent, but much of the research in this field focuses on either the environment \cite{POET, POETenhanced, GenerallyCapable} or the morphology \cite{Scalable, Frank2D}. The research that focuses on both morphology and environment often uses a limited set of environments \cite{Miras, MorphologyComplexity, zhao2020robogrammar}. 

When evolving the morphology and controller simultaneously\cite{lipson2016difficulty}, and when evolving to solve a difficult task directly\cite{scaffolding}, it is common to become stuck in a local optima. Environmental variation could potentially alleviate some of the difficulty by providing stepping stones to more difficult environments, and by introducing environments that require different morphologies to be solved.

One work that considers the optimisation of both agents and environments is the Paired Open-Ended Trailblazer(POET)\cite{POET}. In the field of open-endedness the goal is not to find a single solution, but to find many interesting solutions \cite{OpenEndedness}. In POET agents are optimised to solve environments, at the same time as the environments are optimised for giving the agents new challenges. Constraining the evolving environments by criteria relating to the agent fitness ensures environments that are neither too difficult nor too hard. This creates a push towards novel but solvable environments, leading to an open-ended stream of new tasks.

As mentioned, open-ended algorithms aim to explore as many interesting solutions as possible. Since interestingness is difficult to define \cite{taylor2016open} and optimise for, it is common to take inspiration from novelty search algorithms to instead create solutions that are as different as possible from what has previously been found. The hope is often that finding solutions that are different from each other will make it more likely to find the interesting ones. In novelty search\cite{NoveltySearchOriginal} an archive of previously found solutions is kept, and new solutions are compared to the archive in order to look for solutions that are different from what is already found. This generates a diverse set of solutions. However, in order to make the diverse set of found solutions useful we may also want the solutions to have high quality. This leads us to a family of algorithms called Quality-Diversity algorithms \cite{QDreview, NSLC}, that aim to balance search for novelty with optimisation, to create an archive of solutions that are both diverse and solve their task efficiently. This class of algorithms is often used to ensure that the phenotypic search space is covered, to avoid getting stuck in local optima, while still optimising to solve a set objective.

A popular quality diversity algorithm is MAP-Elites \cite{MAPoriginal}. Map-Elites has been used to optimise both the controller and morphology of robots \cite{MAPmodular, MAPmodularAdaptToDamage}. In MAP-Elites evolution takes place in an archive that is shaped as a grid, where each cell in the grid can hold one solution. MAP-Elites aims to fill the grid while also performing an elitist search for the best candidate within each cell. The position of a candidate within the grid is determined by a set of feature descriptors, called the behaviour dimensions, each relating to one dimension of the grid. These behaviour dimensions are normally set up by the researchers depending on the solution types they want to explore. 

The behaviour dimensions can be difficult to design. In order to avoid creating them by hand, methods for automatic definition of behaviour dimensions have been proposed. AURORA \cite{aurora} uses an autoencoder \cite{Autoencoder}, that has been trained on a set of found candidates, to encode the candidates into a shorter feature vector. This vector determines the placement of the candidates in the MAP-Elites grid. As more candidates are discovered the autoencoder gradually learns a better representation of the search space, allowing the space of possible solutions to be mapped without user-defined descriptors.

Another method for automatically creating behavioural dimensions was presented by Gaier et al.\cite{GaierSkull}, who attempt to use MAP-Elites to reconstruct an image generated by a Compositional pattern producing network (CPPN) \cite{CPPN}. Like AURORA, this method also uses autoencoders. However, instead of using the encoded feature vector produced by the the autoncoder as behaviour dimensions, it instead uses the mean square error between a candidate and the attempted reconstruction of the candidate by the autoencoder. This error value indicates how novel or unexpected the solution is: If the autoencoder cannot reconstruct it, it cannot have seen many similar solutions in its training data. This behavior descriptor thereby becomes a measure of how original a solution is, allowing the exploration of solutions with different degree of familiarity. The autoencoder is retrained at intervals, causing the dimensions to shift to encourage exploration of images different from what is in the map. By keeping solutions with different levels of familiarity, the map gradually builds a record of the most notable images previously explored. As the second behaviour dimension the number of nodes in the CPPN is used.

\begin{figure}
\begin{center}
\includegraphics[width=0.8\textwidth]{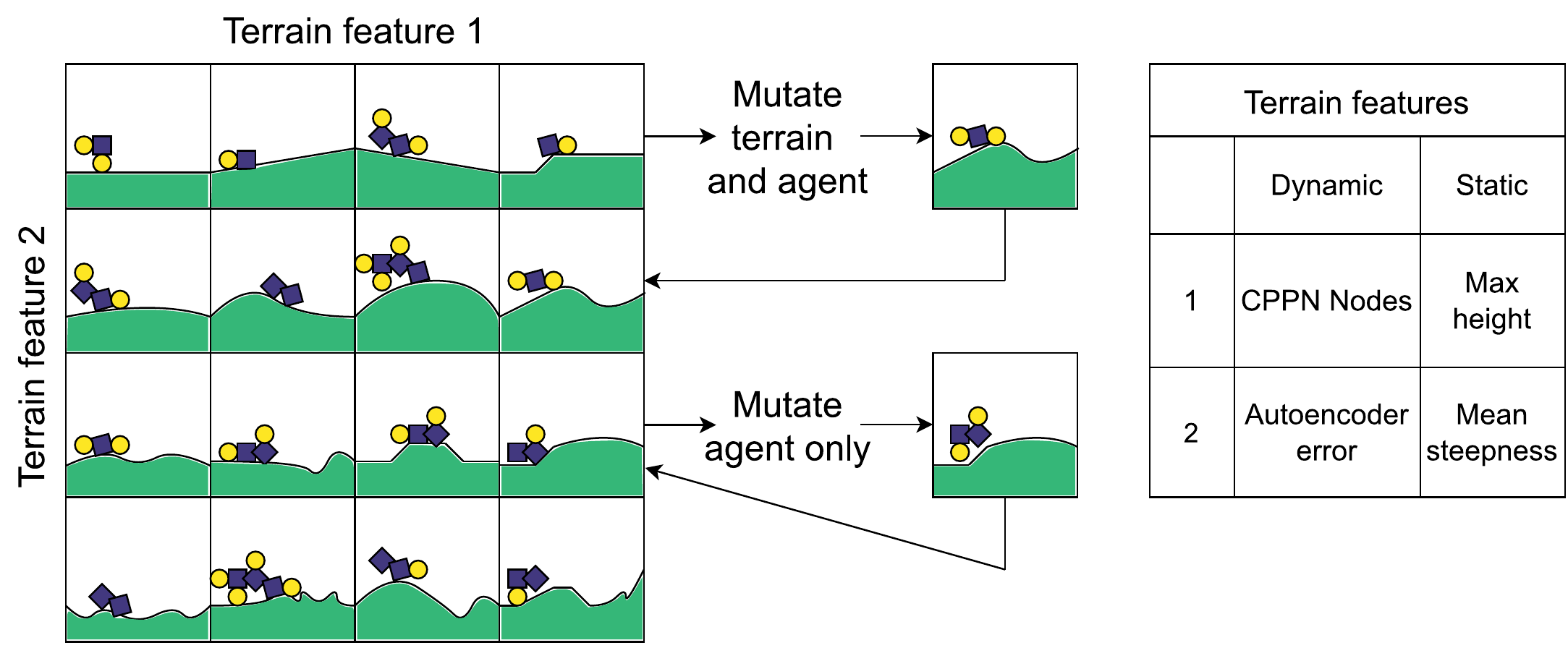}
\end{center}
\caption{Each cell in the map can hold a pair of one terrain and one virtual creature. In each iteration pairs from some cells are chosen to be mutated. The virtual creature is always mutated, while the environment is only mutated for some of the chosen pairs. The mutated pairs are inserted into the map if their fitness is higher than the fitness of the pair already in the cell they belong to.}
\label{fig:concept}
\end{figure}

Our approach takes inspiration from the Paired Open-Ended Trailblazer (POET)\cite{POET, POETenhanced}, which explores pairs of environments and agents solving them, with the goal of endlessly innovating to create ever-more challenging environments and agents solving them. However, unlike POET, our approach attempts a structured exploration of new solutions, aiming to fill up a grid-shaped repertoire of environments, by using the MAP-Elites \cite{MAPoriginal} algorithm (figure \ref{fig:concept}). With this grid structure solutions in adjacent squares can be used as stepping stones for the agents to solve increasingly difficult environments. Our \emph{static} approach uses handcrafted features of the environment as map dimensions. However, to explore open-ended generation of environments we also take inspiration from Gaier \cite{GaierSkull}, and test the use of autoencoder error as a \emph{dynamic} map dimension. We found that while both the static and dynamic approach managed to solve a diverse set of environments, the static bounded dimensions led to larger exploration of environments.

Our contributions are twofold. 1) We explore the possibility of creating a structured repertoire of tasks and agents with MAP-Elites, and show that this approach is capable of generating a diverse set of terrains and virtual creatures that manage to walk through them. In this preliminary work we limit the tasks to locomotion on different terrains. While the terrains are unbounded, the task of locomotion is not. To truly achieve our goal of unlimited tasks we will in the future have to evolve not only the terrain but also the tasks that the agents solve. 2) We test the use of an autoencoder as a behaviour dimension in the map to allow for unbounded innovation.

\section{Methods}
Together the body, brain and environment determines the behaviour of a virtual creature. We evolve all these three components with the goal of finding diverse terrains and agents that walk through them. Inspired by MAP-Elites\cite{MAPoriginal} we optimise within an archive structured as a 2d grid\footnote{Source code is available at \url{https://github.com/EmmaStensby/environment-map}}. Each cell in the grid holds a pair of one agent and one environment. If multiple pairs have been found for a single cell only the pair with the highest fitness score is kept. Terrain features of the environment determines the placement of the pair within the grid. We compare two variants of our approach. The first uses handcrafted grid dimensions, and will be referred to as the static approach. The second uses a combination of handcrafted and automatically defined dimensions, and will be referred to as the dynamic approach. As the testbed for our algorithm we will use a simulation environment created by Veenstra et al.\footnote{\url{https://github.com/FrankVeenstra/gym_rem2D}} \cite{Frank2D}, where 2D modular virtual creatures move through a course, attempting to reach the end. This environment is convenient as it is not computationally heavy, and because it allows changing all three components that we are interested in evolving: Terrain, morphology and controller. 

\subsection{Simulation environment}
We test our approach using an OpenAI Gym\cite{openAI} simulator for 2D modular virtual creatures \cite{Frank2D}, the simulator uses the Box2D physics engine \cite{box2d}. The creatures consist of circles and rectangles, and can be represented as trees. The root module of a creature is always a rectangle. Every rectangle module can connect to up to three new modules. Circle modules cannot connect to any new modules, so all circle modules will be leaf nodes of the virtual creature tree. In addition to their shape the modules have parameters for size and the angle at which they are connected to their parent.

The virtual creature moves across a 2d terrain, which is 220 units long. The first 20 units of the terrain is a startpad, which is always flat. The environment is defined by specifying the height of the terrain at each unit. A creature's fitness is defined as the number of units its root module has progressed along the terrain. The creatures are simulated for up to 2000 time steps, after this the simulation is stopped to ensure that the time spent to evaluate a single individual is not too long. A vertical line moves after the simulated creature at a speed of 0.02 units per time step. If the line reaches the creature the simulation will end, quickly eliminating individuals that do not move.

\subsection{Environment encoding}
Like in POET-Enhanced\cite{POETenhanced}, terrains are generated by a compositional pattern producing network\cite{CPPN} (CPPN). The initialisation and mutation parameters of the CPPN are the same as those used in POET-enhanced, as we wished to use a method for terrain generation already established in the literature. 200 values evenly distributed between 0 and 1 are evaluated by the CPPN to create a vector containing the height of the 200 units of the terrain.

\subsection{Agent encoding}

\begin{table}[]
    \centering
    \begin{tabular}{|l|l|}
        \hline
        \textbf{Module (Rectangle)} & \textbf{12 bits}\\
        \hline
        Bit 0-3 & width \\
        Bit 4-7 & height \\
        Bit 8-11 & angle \\
        \hline
    \end{tabular}
     \begin{tabular}{|l|l|}
        \hline
        \textbf{Module (Circle)} & \textbf{8 bits}\\
        \hline
        Bit 0-3 & radius \\
        Bit 4-7 & angle \\
        - & - \\
        \hline
    \end{tabular}
    \begin{tabular}{|l|l|}
        \hline
        \textbf{Controller} & \textbf{12 bits}\\
        \hline
        Bit 0-3 & amplitude \\
        Bit 4-7 & period \\
        Bit 8-11 & phase \\
        \hline
    \end{tabular}
    \caption{Encoding of a module and controller within the bitstring that encodes a modular virtual creature.}
    \label{tab:encoding}
\end{table}

An agent is encoded as a bitstring, which can be decoded into a tree structure representing both a 2d modular virtual creature and its controller. Representing the agent as a bitstring eases the design of mutation operators. The agents were mutated by flipping bits with a probability of 0.05. We use a decentralised controller where each module is controlled by a sine wave. The bitstring has a length of 288 bits. The first 48 bits are decoded into four rectangle modules, the next 32 bits are decoded into four circle modules, the next 96 bits are decoded into eight controllers. The controller for each module produces a sine wave that controls the angle of a module in relation to its parent module. How the bits relate to the parameters of the modules and controllers is summarised in table \ref{tab:encoding}. The last 112 bits are decoded into a tree where each node contains one of the modules and one of the controllers previously defined. The tree is generated by performing the following steps: 
\begin{enumerate}
    \item Add all available connection points for modules to a list.
    \item Pass through the bitstring until a 1 is reached. For every 0 passed remove one connection point from the top of the list.
    \item The next 6 bits are decoded into two numbers between 0 and 7, which decide which of the eight modules and which of the eight controllers are to be used at the connection point now at the top of the list.
    \item Add all new connections to the list, and repeat from step 2.
\end{enumerate}
These steps continue until the end of the string is reached, or the list of connection points is empty.

\subsection{Environment-Agent MAP-Elites}

\begin{figure}
\centering
\includegraphics[width=0.8\textwidth, interpolate=false]{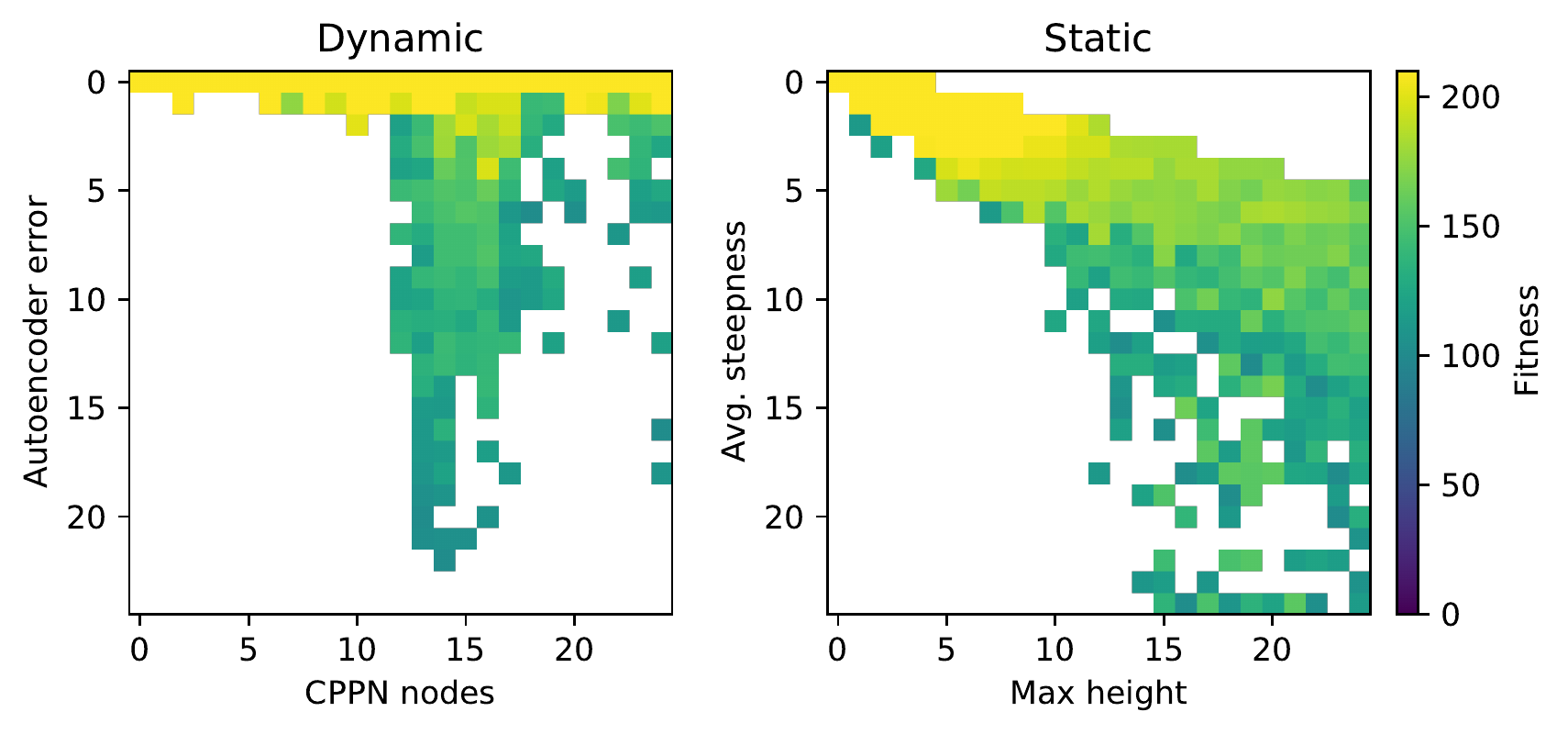}
\caption{Example maps from a dynamic and static run. These maps are the archives used by the MAP-Elites algorithm. The color of each cell represents the fitness of the pair in that position of the grid. The maps are taken from the runs with median \emph{average map fitness} (see fig \ref{fig:boxplot}) out of all 29 runs.}
\label{fig:example_map}
\end{figure}

Our algorithm keeps an archive shaped as a 2d map, the map has 25 by 25 cells. Figure \ref{fig:example_map} shows how the maps may look during runtime. Each cell in the map can hold one environment and one agent. The environment and agent form a pair, and the fitness of the pair is determined the agent's fitness in the environment. The placement of a pair in the map is determined by the map's behaviour dimensions.

Before a run is started the maps are bootstrapped with initial solutions. 500 random pairs of environments and agents are created and placed in their respective cells in the grid. The 500 initial environments are generated by mutating a flat environment. The initial environments will then be spread across a small area of the map. If there are several pairs that belong in the same cell the one with the highest fitness is kept, and the rest are discarded.

Next the MAP-Elites algorithm is applied. One iteration of the algorithm consists of the following three steps:
\begin{description}
    \item[Select] 500 random pairs from the grid, the same pair can be selected multiple times.
    \item[Mutate] the agent in all pairs. Mutate the environment with a probability of 0.2.
    \item[Insert] the mutated pairs into the grid.
\end{description}

When attempting to insert the mutated pairs into the map they are first evaluated to determine their fitness. Next the cell that they belong to is found. If the cell is empty they are inserted as long as their fitness is above a threshold of 100. If the cell is occupied the pair with the highest fitness will be kept, while the other is discarded.

\subsection{Behaviour dimensions}
We compare two different ways of defining the behaviour dimensions of the map, the \emph{static} approach and the \emph{dynamic} approach. The static approach has the maximum height of the terrain as the first dimension, and the average steepness as the second dimension. The dynamic approach has the number of nodes in the CPPN, which generates the terrain, as the first dimension. The second dimension is the reproduction error of an autoencoder. Since the autoencoder is retrained at regular intervals, this dimension changes as the map is filled. The second dimension in both the static and the dynamic approach is scaled by a constant to ensure the map holds reasonable environments. This is necessary because it is difficult to create environments that have very high values for average steepness and autoencoder error. The second dimension for the static approach is scaled by 50, while the second dimension for the dynamic approach is scaled by 5.

The autoencoder used for the dynamic approach has an input layer with 200 nodes, three hidden layers with respectively 64, 32 and 64 nodes, and an output layer with 200 nodes. It is trained with the adam optimiser, and the loss function is the mean squared error. It is bootstrapped by training on 500 randomly generated simple terrains at the start of each run, and is retrained every 100 iterations. When it is retrained it is trained on all environments currently in the map. The reproduction error, used as the novelty measure for the behaviour dimension, is defined as the mean absolute error between the terrain and the reproduced terrain from the autoencoder.

\section{Recording data}

\subsection{Reference maps}
In addition to the map used as the archive for our algorithm we also record the pairs found in a separate map with higher resolution. All explored solutions may be recorded in the reference map, regardless of whether they were placed in the MAP-Elites archive. The reference map has 100 by 100 cells and is used to compare the solutions found by the dynamic approach to those found by the static approach. The behaviour dimensions for the reference maps are the same as the dimensions for the static map, except for the resolution. The static approach therefore has an advantage when filling the reference map as it has access to almost the same dimensions during runtime.

\subsection{Found and solved environments}
We also record explored solutions in two lists. These two archives hold respectively found and solved environments. Each time a new pair is explored is is added to the archive of found environments, as long as there is no environment already in the archive that is too similar to it. An environment is regarded as too similar if there is an environment in the archive to which it has an absolute error of less than 25. If the pair has a fitness above 200 it is also added to the archive of solved environments. The archive for solved environments has a lower threshold for absolute error at 2.5.

\subsection{Environment difficulty}

We analyse some of our results by approximating the environment difficulty. The difficulty is measured by discretising the steepness of the terrain into several categories. Next each hill in the terrain is localised. A hill is defined as a continuous section of terrain units where all units belong to the same steepness category. The hills are then assigned a value based on their length and steepness category, see table \ref{tab:envdiff}. The difficulty of the terrain is the sum of the values for all its hills.

\begin{table}[]
    \centering
    \begin{tabular}{| c l |c | c | c | c | c | c | c|}
        \hline
        & & \multicolumn{6}{c}{Steepness }& \\
        & & $<$-2.4 & $<$-0.24 & $<$-0.024 & -0.024 to 0.024 & $>$0.024 & $>$0.24 & $>$2.4 \\
        \hline
        & 1-3 & -3  & -2 & -1 & 0 & 2 & 4 & 6\\
        \rotatebox[origin=c]{90}{Units} & 4-8 & -4 & -3 & -2 & 0 & 4 & 6 & 8\\
        & $>$8 & -5 & -4 & -3 & 0 & 6 & 8 & 10\\
        \hline
    \end{tabular}
    \caption{Difficulty values for hills.}
    \label{tab:envdiff}
    \vspace{-0.5cm} 
\end{table}

\section{Results}

To select parameters 200 trials with random parameters were performed for each of the two approaches. The parameters were handpicked based on the results of the trials. Next we performed 29 runs of each of our two approaches on 16 cores for 16 hours.
The number of iterations completed within this time varied between the runs. However, the average number of iterations was 1364 which corresponds to evaluating  682 000 individuals. 

\subsection{Reference maps and performance}

\begin{figure}
\includegraphics[width=\textwidth, interpolate=false]{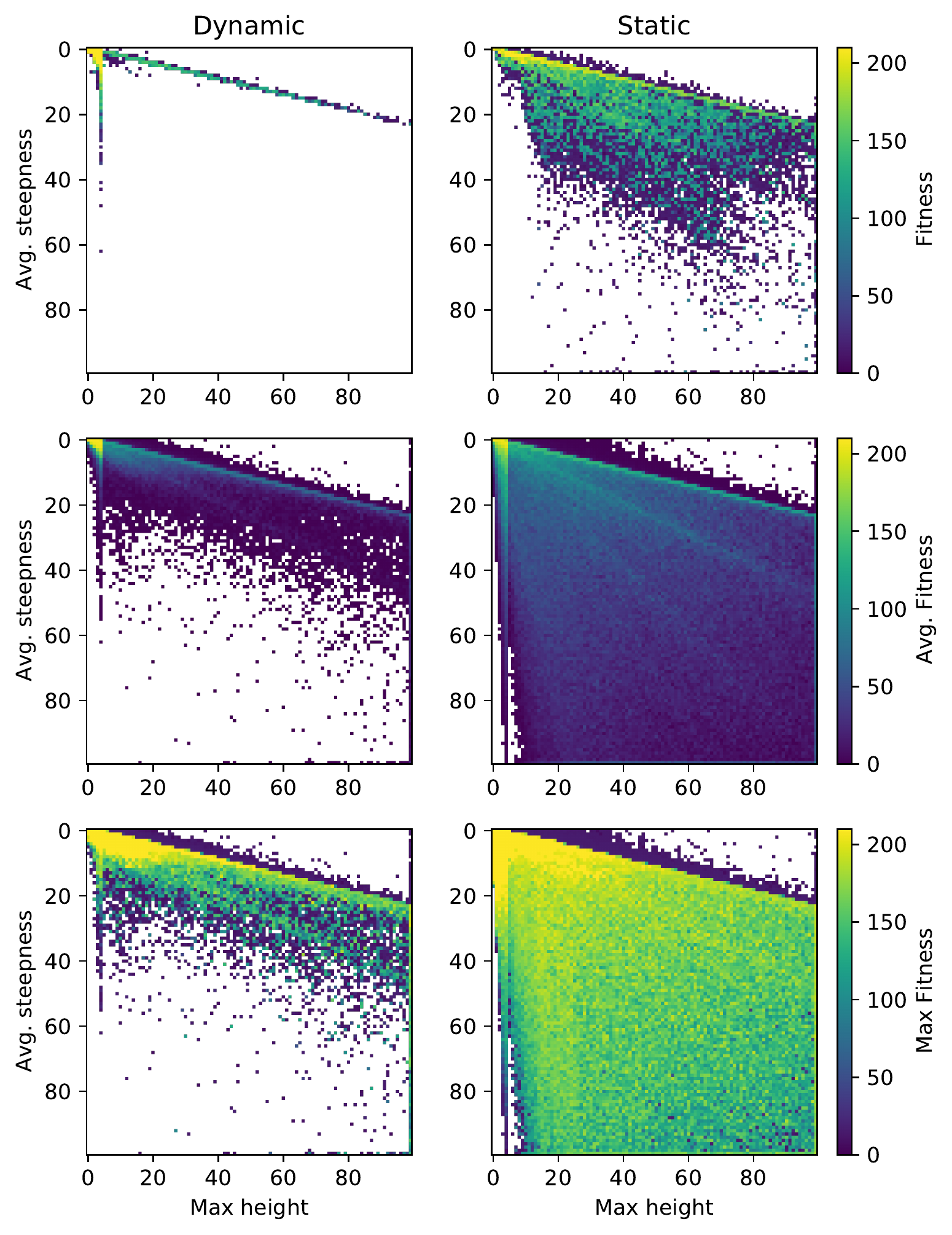}
\caption{Reference maps record the progress of the two approaches. The color of each square represents the fitness of the pair found in that cell. The top row shows the reference maps from single example runs. The example maps are taken from the runs with median \emph{average map fitness} out of all 29 runs. The second row shows the mean fitness, and the bottom row shows the maximum fitness over all 29 runs.}
\label{fig:all_map}
\end{figure}

In figure \ref{fig:all_map} we see the reference maps for the two approaches. We can see that the dynamic approach has explored a significantly smaller part of the map than the static approach. This is expected as the static approach optimises directly to fill dimensions very similar to those of the reference map. Both of the approaches has high fitness in environments in the top left corner of the map, and decreasing fitness towards the bottom right.
In the top right corner there is an area where all found pairs have 0 fitness. In this area it is not possible to create solvable environments due to the terrain features required by the map dimensions. These environments have high maximum terrain height, but low average steepness. This is only possible to achieve by setting the first step in the terrain to a high value, which creates a tall wall immediately after the startpad. 

\begin{figure}
\centering
\includegraphics[width=0.8\textwidth]{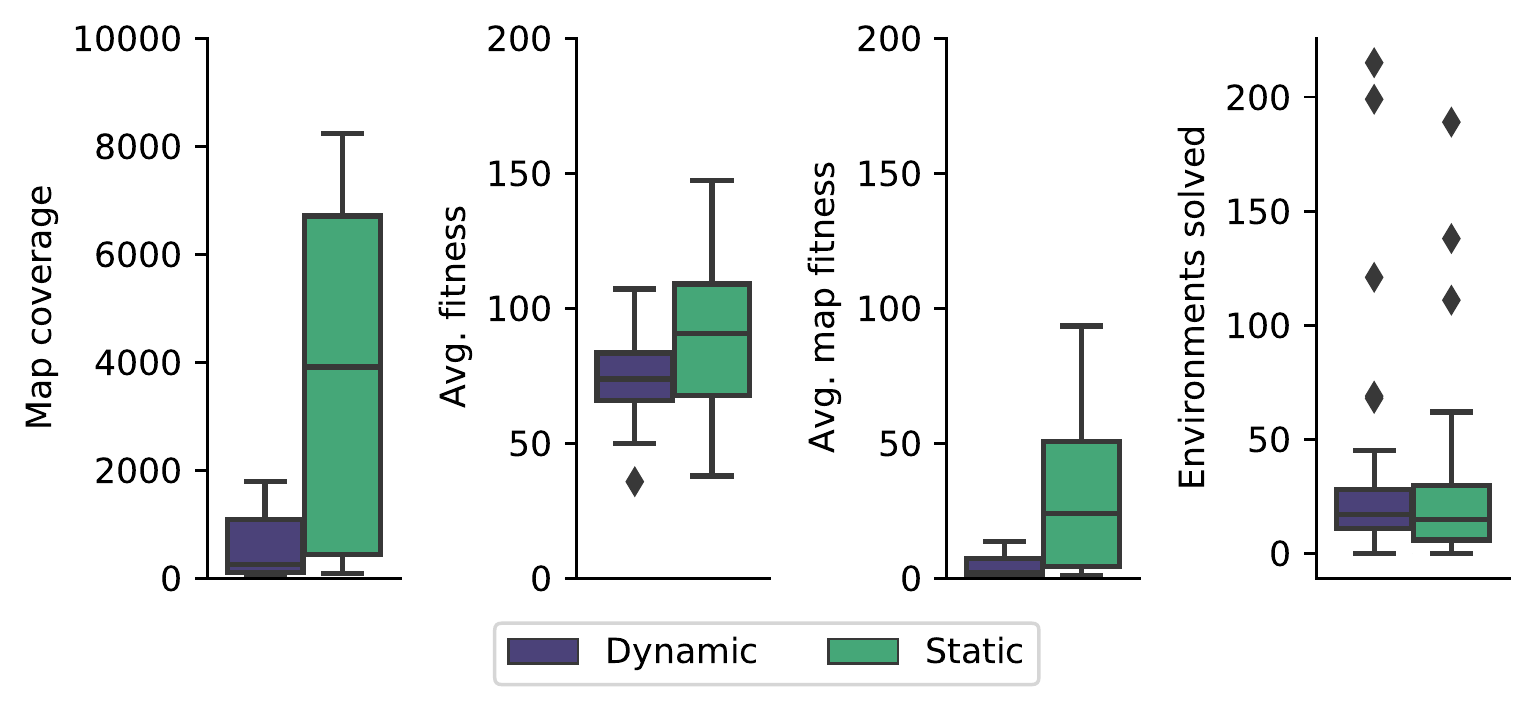}
\caption{From left to right we see 1) the coverage of the reference maps, 2) the average fitness of the pairs present in the reference maps 3) the sum of the fitness of all individuals in the reference maps divided by the number of squares, and 4) the number of environments solved.}
\label{fig:boxplot}
\end{figure}

Figure \ref{fig:boxplot} shows statistics about the performance of the 29 runs. The static approach performs better than the dynamic approach in both coverage of the reference map, and average map fitness. The average map fitness is the average fitness per square in the reference map. The two approaches performed similarly for average fitness of the found solutions and the total number of solved environments. 

\begin{figure}
\centering
\includegraphics[width=0.8\textwidth]{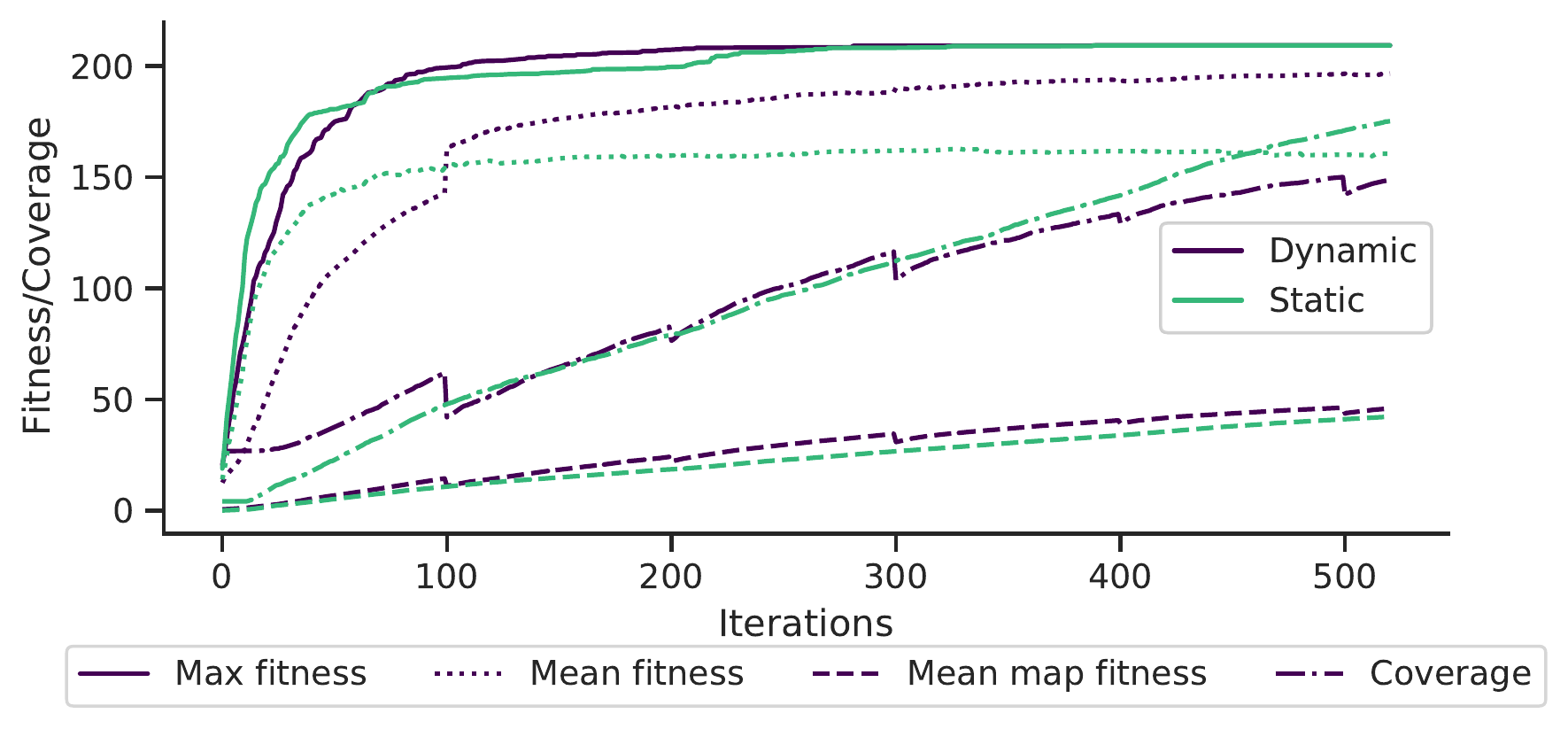}
\caption{This graph shows 1) maximum fitness, 1) mean fitness of found pairs, 3) average map fitness and 4) map coverage over time. These are measured on the maps used as archives by MAP-Elites during runtime (see figure \ref{fig:example_map}). The graphs show the mean over all 29 runs. Note that this figure excludes some runs that completed very few iterations and is therefore meant only to illustrate the effect of the autoencoder training on the behaviour dimensions.}
\label{fig:graph}
\end{figure}

In figure \ref{fig:graph} we can see how the coverage of the MAP-Elites archives develops for both approaches. This graph excludes some runs that completed very few iterations and is therefore meant only to illustrate the effect of the autoencoder training on the behaviour dimensions. For the dynamic approach we can see the effect of the autoencoder training every 100 iterations. As the behaviour dimensions change, some pairs that were previously in separate cells end up in the same cell, and the coverage drops slightly.

\subsection{Analysis of found environments}

\begin{figure}
\centering
\includegraphics[width=0.8\textwidth]{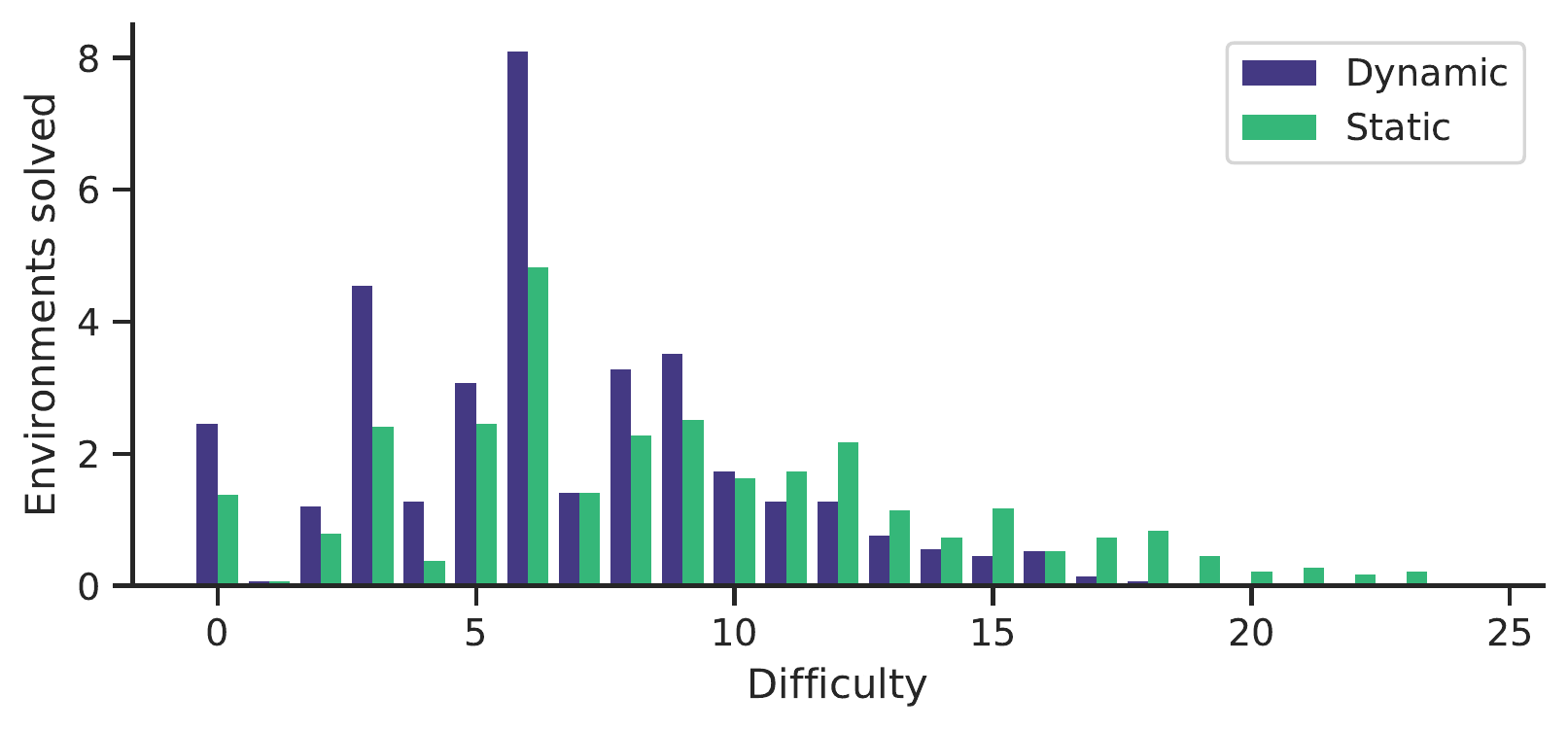}
\caption{This histogram shows the distribution in difficulty of the solved environments. Each bar shows the mean number of environments solved for that difficulty over all 29 runs.}
\label{fig:solved}
\end{figure}

In figure \ref{fig:solved} we can see how the solved environments are distributed with regards to difficulty. The distribution of the solved environments is slightly different for the two approaches. Although the two approaches has solved approximately the same number of environments, the dynamic approach seems to have solved more simple environments, while the static approach has solved quite a few difficult environments.

\begin{figure}
\centering
\includegraphics[width=0.8\textwidth, interpolate=false]{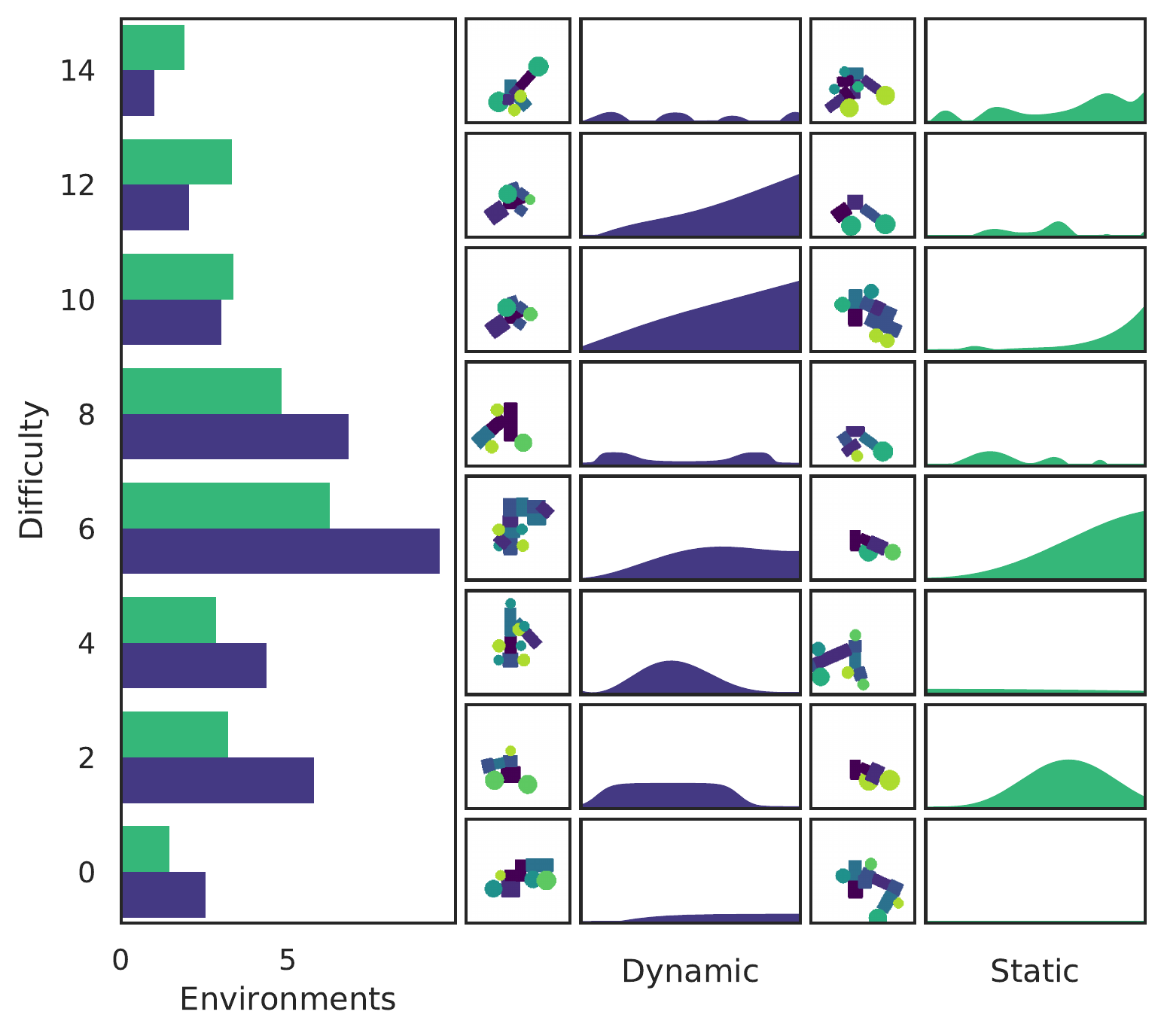}
\caption{The histogram on the left shows the same data as the histogram in figure \ref{fig:solved}. However, the number of bins have been halved by combining every two bins. On the right a pair from each bin is shown. The pairs are drawn randomly from their respective bins. Note that the virtual creatures shown are larger than their actual size compared to the environments, and that the y axis for the environments has been scaled to highlight terrain features. However, all the displayed environments are scaled equally so they can be compared to each other.}
\label{fig:courses}
\end{figure}

Figure \ref{fig:courses} displays some examples of solved environments. The environments have been scaled to highlight terrain features.
We can see qualitatively that the algorithm produces various different terrains and agents.

\begin{figure}
\centering
\includegraphics[width=0.8\textwidth]{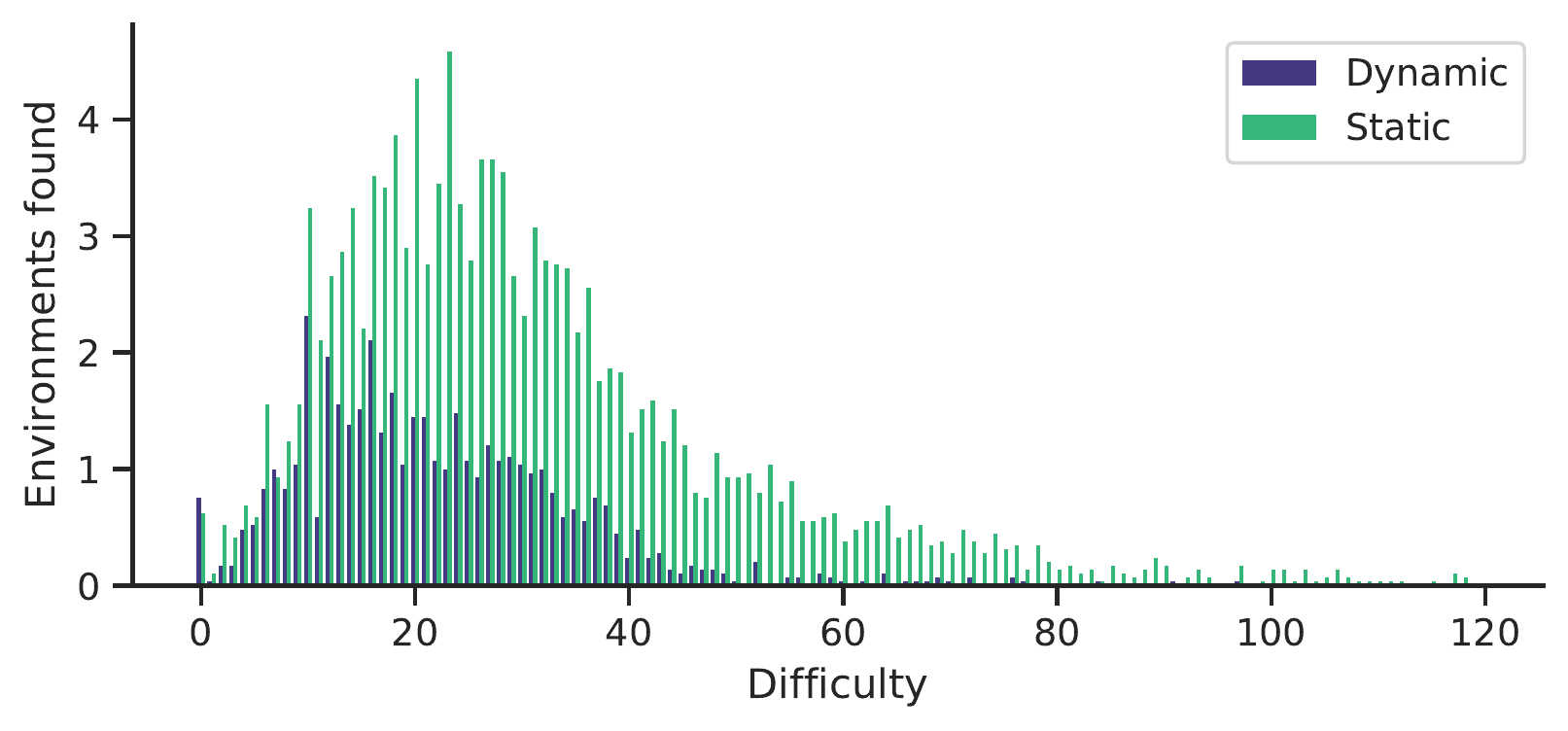}
\caption{This histogram shows the distribution in difficulty of the found, but not necessarily solved, environments. Each bar shows the mean number of environments found for that difficulty over all 29 runs.}
\label{fig:found}
\end{figure}

In figure \ref{fig:found} we quantitatively analyse all explored environments with regards to difficulty. We can clearly see that the static approach has explored significantly more environments than the dynamic approach.

\section{Discussion}
We have explored the possibility of creating a structured repertoire of environments, and agents solving them, using MAP-Elites. We expected the grid structure in MAP-Elites to aid the agents in exploring and solving increasingly difficult environments by using adjacent squares in the map as stepping stones. The reference maps in figure \ref{fig:all_map} showed us that our approach is indeed capable of filling a map with environments and agents. The environments seem to be distributed as expected within the map, with the easy environments in the top left corner, and difficult environments in the bottom right corner. This is verified by the fitness found gradually decreasing towards the bottom right. We can see qualitatively in figure \ref{fig:courses} that the approach seems capable of solving diverse environments, as the randomly drawn solved environments are quite different from each other. Although no conclusions can be drawn from the few environments plotted, they seem to get increasingly bumpy as the difficulty increases.

The dynamic approach filled significantly less of the reference maps in figure~\ref{fig:all_map} than the static approach. This was an expected result as the static approach optimises directly to explore the features of the reference map, while the dynamic approach optimises for a different novelty measure. Figure \ref{fig:found} showed a quantitative analysis of the environments found, and confirms that the static approach explored a larger diversity of environments. While handcrafted behaviour dimensions may be difficult to create in some cases, they performed better than the automatically defined dimensions in our case. However, in other, more complex, domains where handcrafted dimensions may be more difficult to create, the automatically defined dimensions from the autoencoder can be an alternative that is more general and can be applied to most types of environments. 

Another benefit of the dynamic approach is that it could in theory continue exploring new environments for longer than the static approach. The static approach would likely stagnate once all squares in its map have been filled with high fitness pairs that are difficult to replace, while the dynamic approach could keep retraining the autoencoder and change its dimensions. We did not have the opportunity to see whether such an effect would appear in our experiments, as we did not run the experiment for long enough for this to happen. A main limitation for such continuous exploration is the capability of the autoencoder. A requirement for the dynamic approach to keep endlessly exploring the available environments in more detail, is that the autoencoder is capable of storing information about all found environments and discern new environments from these. This becomes increasingly difficult as more environments are discovered, and further experiments may be necessary for the autoencoder to be able to do its part.

We do not only want to \emph{generate} diverse terrains, we also want to \emph{solve} them. 
We discovered that the static and dynamic approach had solved approximately the same number of environments (figure~\ref{fig:solved}), despite the difference in the number of found environments (figure~\ref{fig:found}). 
This may indicate that although the dynamic approach has explored less of the environment search space, it may have explored the top left section of the maps more thoroughly, leading to the increased number of solved easy environments. 

\section{Conclusion and future work}
This work was an attempt to explore the potential of using MAP-Elites to generate both interesting tasks and their solutions. We found that a map with handcrafted bounded dimensions lead to the exploration of a large set of environments. We compared bounded and unbounded behaviour dimensions, and both approaches managed to solve a diverse set of environments.
The main limitation of our approach seems to be that it is challenging to create general and open-ended behaviour dimensions for the map, that actually allow for endless generation of new environments. 
For the map to be able to continually develop novel tasks, the autoencoder used to describe the novelty must be able to represent many previously found environments, and meaningfully discern them from new environments. It would be interesting to further explore the choice of behaviour dimensions. Either by finding out what properties are necessary for the autoencoder to perform well, even as the amount of data it trains on becomes very large, or by exploring alternative behaviour dimensions. An approach like AURORA \cite{aurora} could be explored as an alternative to our approach, as its feature extraction may lead to an interesting structure to the repertoire.

Another direction could be to explore possible uses of an already generated repertoire of environments and solutions. While generating novel environments automatically is interesting in itself, it could be even more interesting if the skills stored in the repertoire could somehow be leveraged to quickly adapt to new never before seen environments. In this case it would be interesting to either design the behaviour dimensions of the map so that they correlate with the skills necessary to solve the environments, or use methods for adapting through trial and error~\cite{AdaptToDamage}. 

Other extensions to our current work could include improving the efficiency of the search within the map, for example by introducing crossover or other mechanisms that create interaction throughout the map. The efficiency should also be compared with other existing methods that generate terrains and their solutions, such as the Paired Open-Ended Trailblazer \cite{POETenhanced}, or to quality-diversity methods with unstructured repertoires, such as Novelty Search with Local Competition~\cite{NSLC}.

\section*{Acknowledgments}
This work was partially supported by the Research Council of Norway through its Centres of Excellence scheme, project number 262762. The simulations were performed on resources provided by UNINETT Sigma2 - the National Infrastructure for High Performance Computing and Data Storage in Norway. Thank you to Frank Veenstra for support using the 2D simulator for modular robots.

\printbibliography

@article{Autoencoder,
  title={Reducing the dimensionality of data with neural networks},
  author={Hinton, Geoffrey E and Salakhutdinov, Ruslan R},
  journal={science},
  volume={313},
  number={5786},
  pages={504--507},
  year={2006},
  publisher={American Association for the Advancement of Science}
}

@misc{openAI,
  Author = {Greg Brockman and Vicki Cheung and Ludwig Pettersson and Jonas Schneider and John Schulman and Jie Tang and Wojciech Zaremba},
  Title = {OpenAI Gym},
  Year = {2016},
  Eprint = {arXiv:1606.01540},
}

@misc{box2d,
  Author = {Erin Catto},
  Title = {Box2D},
  Year = {2019},
  Eprint = {https://box2d.org/},
}

@article{CPPN,
  title={Compositional pattern producing networks: A novel abstraction of development},
  author={Stanley, Kenneth O},
  journal={Genetic programming and evolvable machines},
  volume={8},
  number={2},
  pages={131--162},
  year={2007},
  publisher={Springer}
}

@article{Frank2D,
  author = {Veenstra, Frank and Glette, Kyrre},
  title = {How Different Encodings Affect Performance and Diversification when Evolving the Morphology and Control of 2D Virtual Creatures},
  journal = {Artificial Life Conference Proceedings},
  volume = {},
  number = {32},
  pages = {592-601},
  year = {2020},
  doi = {10.1162/isal\_a\_00295},
  URL = { https://www.mitpressjournals.org/doi/abs/10.1162/isal_a_00295 },
  eprint = { https://www.mitpressjournals.org/doi/pdf/10.1162/isal_a_00295 }
}

@article{taylor2016open,
  title={Open-ended evolution: Perspectives from the OEE workshop in York},
  author={Taylor, Tim and Bedau, Mark and Channon, Alastair and Ackley, David and Banzhaf, Wolfgang and Beslon, Guillaume and Dolson, Emily and Froese, Tom and Hickinbotham, Simon and Ikegami, Takashi and others},
  journal={Artificial life},
  volume={22},
  number={3},
  pages={408--423},
  year={2016},
  publisher={MIT Press}
}

@article{OpenEndedness,
  author = {Stanley, Kenneth O.},
  title = {Why Open-Endedness Matters},
  year = {2019},
  issue_date = {Summer 2019},
  publisher = {MIT Press},
  address = {Cambridge, MA, USA},
  volume = {25},
  number = {3},
  issn = {1064-5462},
  url = {https://doi.org/10.1162/artl_a_00294},
  doi = {10.1162/artl_a_00294},
  journal = {Artificial Life},
  month = {aug},
  pages = {232–235},
  numpages = {4}
}

@ARTICLE{NoveltySearchOriginal,
  author={Lehman, Joel and Stanley, Kenneth O.},
  journal={Evolutionary Computation}, 
  title={Abandoning Objectives: Evolution Through the Search for Novelty Alone}, 
  year={2011},
  volume={19},
  number={2},
  pages={189-223},
  doi={10.1162/EVCO_a_00025}
}

@inproceedings{NSLC,
author = {Lehman, Joel and Stanley, Kenneth O.},
title = {Evolving a Diversity of Virtual Creatures through Novelty Search and Local Competition},
year = {2011},
isbn = {9781450305570},
publisher = {Association for Computing Machinery},
address = {New York, NY, USA},
url = {https://doi.org/10.1145/2001576.2001606},
doi = {10.1145/2001576.2001606},
booktitle = {Proceedings of the 13th Annual Conference on Genetic and Evolutionary Computation},
pages = {211–218},
numpages = {8},
location = {Dublin, Ireland},
series = {GECCO '11}
}

@inproceedings{GaierSkull,
author = {Gaier, Adam and Asteroth, Alexander and Mouret, Jean-Baptiste},
title = {Are Quality Diversity Algorithms Better at Generating Stepping Stones than Objective-Based Search?},
year = {2019},
isbn = {9781450367486},
publisher = {Association for Computing Machinery},
address = {New York, NY, USA},
url = {https://doi.org/10.1145/3319619.3321897},
doi = {10.1145/3319619.3321897},
booktitle = {Proceedings of the Genetic and Evolutionary Computation Conference Companion},
pages = {115–116},
numpages = {2},
location = {Prague, Czech Republic},
series = {GECCO '19}
}

@article{MAPoriginal,
  title={Illuminating search spaces by mapping elites},
  author={Mouret, Jean-Baptiste and Clune, Jeff},
  journal={arXiv preprint arXiv:1504.04909},
  year={2015}
}

@ARTICLE{MAPmodular,
AUTHOR={Nordmoen, Jørgen and Veenstra, Frank and Ellefsen, Kai Olav and Glette, Kyrre},   
TITLE={MAP-Elites Enables Powerful Stepping Stones and Diversity for Modular Robotics},      
JOURNAL={Frontiers in Robotics and AI},      
VOLUME={8},      
PAGES={56},     
YEAR={2021},      
URL={https://www.frontiersin.org/article/10.3389/frobt.2021.639173},
DOI={10.3389/frobt.2021.639173},
ISSN={2296-9144}
}

@inproceedings{MAPmodularAdaptToDamage,
  title={Learning behaviour-performance maps with meta-evolution},
  author={Bossens, David M and Mouret, Jean-Baptiste and Tarapore, Danesh},
  booktitle={Proceedings of the 2020 Genetic and Evolutionary Computation Conference},
  pages={49--57},
  year={2020}
}

@incollection{QDreview,
  title={Quality-Diversity Optimization: a novel branch of stochastic optimization},
  author={Chatzilygeroudis, Konstantinos and Cully, Antoine and Vassiliades, Vassilis and Mouret, Jean-Baptiste},
  booktitle={Black Box Optimization, Machine Learning, and No-Free Lunch Theorems},
  pages={109--135},
  year={2021},
  publisher={Springer}
}

@article{Miras,
  title={Environmental influences on evolvable robots},
  author={Miras, Karine and Ferrante, Eliseo and Eiben, Agoston E},
  journal={PloS one},
  volume={15},
  number={5},
  pages={e0233848},
  year={2020},
  publisher={Public Library of Science San Francisco, CA USA}
}

@article{MorphologyComplexity,
  title={Environmental influence on the evolution of morphological complexity in machines},
  author={Auerbach, Joshua E and Bongard, Josh C},
  journal={PLoS computational biology},
  volume={10},
  number={1},
  pages={e1003399},
  year={2014},
  publisher={Public Library of Science San Francisco, USA}
}

@article{Scalable,
  title={Scalable co-optimization of morphology and control in embodied machines},
  author={Cheney, Nick and Bongard, Josh and SunSpiral, Vytas and Lipson, Hod},
  journal={Journal of The Royal Society Interface},
  volume={15},
  number={143},
  pages={20170937},
  year={2018},
  publisher={The Royal Society}
}

@inproceedings{scaffolding,
author = {Bongard, Josh C.},
title = {Morphological and Environmental Scaffolding Synergize When Evolving Robot Controllers: Artificial Life/Robotics/Evolvable Hardware},
year = {2011},
isbn = {9781450305570},
publisher = {Association for Computing Machinery},
address = {New York, NY, USA},
url = {https://doi.org/10.1145/2001576.2001602},
doi = {10.1145/2001576.2001602},
booktitle = {Proceedings of the 13th Annual Conference on Genetic and Evolutionary Computation},
pages = {179–186},
numpages = {8},
location = {Dublin, Ireland},
series = {GECCO '11}
}

@inproceedings{lipson2016difficulty,
  title={On the difficulty of co-optimizing morphology and control in evolved virtual creatures},
  author={Lipson, Hod and Sunspiral, Vytas and Bongard, Josh and Cheney, Nicholas},
  booktitle={Artificial Life Conference Proceedings 13},
  pages={226--233},
  year={2016},
  organization={MIT Press}
}

@article{zhao2020robogrammar,
  title={RoboGrammar: Graph Grammar for Terrain-Optimized Robot Design},
  author={Zhao, Allan and Xu, Jie and Konaković Luković, Mina and Hughes, Josephine and Speilberg, Andrew and Rus, Daniela and Matusik, Wojciech},
  journal={ACM Transactions on Graphics (TOG)},
  volume={39},
  number={6},
  pages={1--16},
  year={2020},
  publisher={ACM New York, NY, USA}
}

@article{AdaptToDamage,
  title={Robots that can adapt like animals},
  author={Cully, Antoine and Clune, Jeff and Tarapore, Danesh and Mouret, Jean-Baptiste},
  journal={Nature},
  volume={521},
  number={7553},
  pages={503--507},
  year={2015},
  publisher={Nature Publishing Group}
}

@article{GenerallyCapable,
  title={Open-ended learning leads to generally capable agents},
  author={Team, Open Ended Learning and Stooke, Adam and Mahajan, Anuj and Barros, Catarina and Deck, Charlie and Bauer, Jakob and Sygnowski, Jakub and Trebacz, Maja and Jaderberg, Max and Mathieu, Michael and others},
  journal={arXiv preprint arXiv:2107.12808},
  year={2021}
}

@inproceedings{POET,
  author = {Wang, Rui and Lehman, Joel and Clune, Jeff and Stanley, Kenneth O.},
  title = {POET: Open-Ended Coevolution of Environments and Their Optimized Solutions},
  year = {2019},
  isbn = {9781450361118},
  publisher = {Association for Computing Machinery},
  address = {New York, NY, USA},
  url = {https://doi.org/10.1145/3321707.3321799},
  doi = {10.1145/3321707.3321799},
  booktitle = {Proceedings of the Genetic and Evolutionary Computation Conference},
  pages = {142–151},
  numpages = {10},
  location = {Prague, Czech Republic},
  series = {GECCO '19}
}

@inproceedings{POETenhanced,
  title={Enhanced POET: Open-ended reinforcement learning through unbounded invention of learning challenges and their solutions},
  author={Wang, Rui and Lehman, Joel and Rawal, Aditya and Zhi, Jiale and Li, Yulun and Clune, Jeffrey and Stanley, Kenneth},
  booktitle={International Conference on Machine Learning},
  pages={9940--9951},
  year={2020},
  organization={PMLR}
}

@inproceedings{aurora,
author = {Cully, Antoine},
title = {Autonomous Skill Discovery with Quality-Diversity and Unsupervised Descriptors},
year = {2019},
isbn = {9781450361118},
publisher = {Association for Computing Machinery},
address = {New York, NY, USA},
url = {https://doi.org/10.1145/3321707.3321804},
doi = {10.1145/3321707.3321804},
booktitle = {Proceedings of the Genetic and Evolutionary Computation Conference},
pages = {81–89},
numpages = {9},
location = {Prague, Czech Republic},
series = {GECCO '19}
}

\end{document}